\documentclass[conference]{IEEEtran}
\IEEEoverridecommandlockouts
\usepackage{cite}
\usepackage{amsmath,amssymb,amsfonts}
\usepackage{algorithmic}
\usepackage{graphicx}
\usepackage{textcomp}
\usepackage{xcolor}
\usepackage{xurl}
\usepackage{multirow}
\usepackage{makecell}
\def\BibTeX{{\rm B\kern-.05em{\sc i\kern-.025em b}\kern-.08em
    T\kern-.1667em\lower.7ex\hbox{E}\kern-.125emX}}

\begin{document}

\title{Mitigating Algorithmic Bias in Multiclass CNN Classifications Using Causal Modeling
}

\author{\IEEEauthorblockN{1\textsuperscript{st} Min Sik Byun}
\IEEEauthorblockA{\textit{InfoComm Custer} \\
\textit{Singapore Institute of Technology}\\
Singapore \\
2102150@sit.singaporetech.edu.sg}
\and
\IEEEauthorblockN{2\textsuperscript{nd} Wendy Wan Yee Hui}
\IEEEauthorblockA{\textit{InfoComm Custer} \\
\textit{Singapore Institute of Technology}\\
Singapore \\
wendy.hui@singaporetech.edu.sg}
\and
\IEEEauthorblockN{3\textsuperscript{rd} Wai Kwong Lau}
\IEEEauthorblockA{\textit{Sch. of Physics, Mathematics and Computing} \\
\textit{University of Western Australia}\\
Australia \\
john.lau@uwa.edu.au}
}

\maketitle
\thispagestyle{plain}
\pagestyle{plain}

\begin{abstract}
This study describes a procedure for applying causal modeling to detect and mitigate algorithmic bias in a multiclass classification problem. The dataset was derived from the FairFace dataset, supplemented with emotional labels generated by the DeepFace pre-trained model. A custom Convolutional Neural Network (CNN) was developed, consisting of four convolutional blocks, followed by fully connected layers and dropout layers to mitigate overfitting. Gender bias was identified in the CNN model’s classifications: Females were more likely to be classified as “happy” or “sad,” while males were more likely to be classified as “neutral.” To address this, the one-vs-all (OvA) technique was applied. A causal model was constructed for each emotion class to adjust the CNN model's predicted class probabilities. The adjusted probabilities for the various classes were then aggregated by selecting the class with the highest probability. The resulting debiased classifications demonstrated enhanced gender fairness across all classes, with negligible impact—or even a slight improvement—on overall accuracy. This study highlights that algorithmic fairness and accuracy are not necessarily trade-offs. All data and code for this study are publicly available for download.
\end{abstract}


\section{Introduction}

Artificial Intelligence (AI) has transformed decision-making in various domains such as healthcare, finance, and law enforcement. While AI enhances efficiency, a growing concern revolves around its fairness and ethical implications \cite{b1}, \cite{b2}. Central to this concern is the question: Is AI truly fair? The answer to this question depends on one’s worldview. According to \cite{b3}, the “We’re All Equal” (WAE) worldview posits that bias may exist in the dataset, whereas the “What You See Is What You Get” (WYSIWYG) worldview assumes that observations reflect the truth. For example, a researcher holding the WAE worldview would argue that performance differences observed across groups are attributable to structural biases in society. In contrast, a researcher holding the WYSIWYG worldview might contend that it is reasonable for factors affecting performance (e.g., physical strength) to correlate with demographic characteristics. It is also arguable that one may adopt different worldviews depending on the circumstances.

In the context of AI fairness, sensitive data, such as gender, race, age and religion, are referred to as protected attributes \cite{b1}. A commonly used metric to evaluate AI fairness is demographic parity, which requires the AI output to maintain an acceptable level of disparity between protected and non-protected groups. While this aligns closely with the WAE worldview, demographic parity does not account for potential correlations between  protected attributes and unprotected attributes. This limitation can lead to outcomes that some might perceive as “reverse discrimination” from the WYSIWYG perspective. For example, in the context of job hiring, demographic parity could result in the acceptance of unqualified individuals from protected groups to ensure demographic parity \cite{b4}. Mathematically, achieving fairness from the WYSIWYG perspective is a more complex challenge, and it is the primary focus of this study.

\cite{b5} demonstrated the use of causal modeling as a simple tool for detecting and mitigating algorithmic bias. However, their work focused solely on binary classification. In real-life scenarios, many classification problems involve multiple classes. This paper aims to extend their work by applying causal modeling to a multiclass problem. Similar to \cite{b4}, \cite{b5}, and \cite{b6}, we adopt a black-box, post-processing approach for algorithmic bias mitigation. While studies on algorithmic biases have predominantly focused on traditional regression-based algorithms \cite{b5}, \cite{b7}, leaving gaps in understanding the applicability of causal modeling to deep learning models, this paper applies causal modeling to the outputs of convolutional neural networks (CNNs), which have recently gained popularity in image recognition and classification tasks.

In the rest of this paper, we will review related studies, present our methodology and analysis, discuss the findings, and identify some future research directions.

\section{Related Work}

The three general approaches to mitigating algorithmic bias are: (1) pre-processing, (2) in-processing, and (3) post-processing. Pre-processing involves manipulating data before training the AI model. Techniques for pre-processing include re-sampling, re-weighting, altering data representations to balance demographic distributions, and domain adaptation. For example, \cite{b8} proposed data augmentation methods to generate synthetic samples, addressing demographic imbalances and mitigating bias while preserving model accuracy. \cite{b9} applied data augmentation and domain adaptation techniques to improve fairness in detecting age-related macular degeneration (AMD), leading to more accurate and equitable outcomes.

In-processing approaches integrate fairness into the model training process by directly modifying the learning algorithm. For example, \cite{b10} incorporated fairness constraints into their machine learning algorithm. \cite{b11} adopted adversarial training to reduce gender bias in a binary classification problem and demonstrated that a large amount of data is not always necessary to enhance fairness. Hybrid methods that combine pre-processing and in-processing techniques have also shown promise. For instance, generative models have been used to address data imbalances while simultaneously minimizing dependencies between protected attributes and model predictions \cite{b12}. While in-processing methods often yield promising fairness outcomes, they tend to be more computationally intensive and require significant modifications to standard training procedures.

Post-processing focuses on mitigating bias after the model has been trained, treating the model as a black box. This approach involves adjusting the model's outputs to align with fairness metrics, such as equalized odds or equal opportunity. While \cite{b4} ensured equitable predictive outcomes across demographic groups in binary classification, \cite{b6} extended the post-processing techniques to multiclass classification tasks. Although post-processing is flexible and computationally efficient, it cannot be applied in cases where protected attributes are unavailable at inference time \cite{b11}.

Different approaches to bias mitigation have their own strengths and weaknesses, and this study focuses on post-processing. Our method utilizes causal modeling \cite{b13}, \cite{b14}, which is relatively new in the study of AI fairness. Causal models provide a structured framework for identifying and addressing bias by examining the relationships between features, protected attributes, and predictions. Additionally, causal models allow us to evaluate counterfactual fairness \cite{b15}, which requires a decision to remain the same in a counterfactual world where the individual in question belongs to a different demographic group. \cite{b16} demonstrated the potential of causal modeling to improve prediction accuracy in the presence of confounding factors using health datasets. Similarly, \cite{b17} used causal models to identify and quantify biases in both synthetic and real-world datasets. These studies highlight the potential of causal models for algorithmic bias detection. \cite{b5} and \cite{b7} extended the application of causal models to both bias detection and mitigation, with a focus on traditional regression algorithms. However, the effectiveness of the causal modeling approach has not yet been demonstrated in complex architectures such as CNNs. Nor has the approach been applied to multiclass classification problems. To address these gaps, we apply causal modeling to mitigate bias in a multiclass classification CNN model. We aim to address biases in neural network predictions without sacrificing performance. 

\section{Methodology}

The dataset and the CNN model described in Sections \ref{sec3_1} and \ref{sec3_2} were prepared using Python, and the code is available on Google Colab\footnote{\url{https://colab.research.google.com/drive/1CxRZ7QxX2b09ee0pF3GbwA0lZ_syooeQ?usp=sharing}}. The causal model for bias detection and mitigation described in Section \ref{sec3_3} was implemented using R, and the code is also available on Google Colab\footnote{\url{https://colab.research.google.com/drive/1waP5buBpaMPB2Ne5WC9yKTzrMuM0WWkn?usp=sharing}}.

\subsection{Data}\label{sec3_1}

\begin{figure}[hbtp]
\centerline{\includegraphics[width=\linewidth]{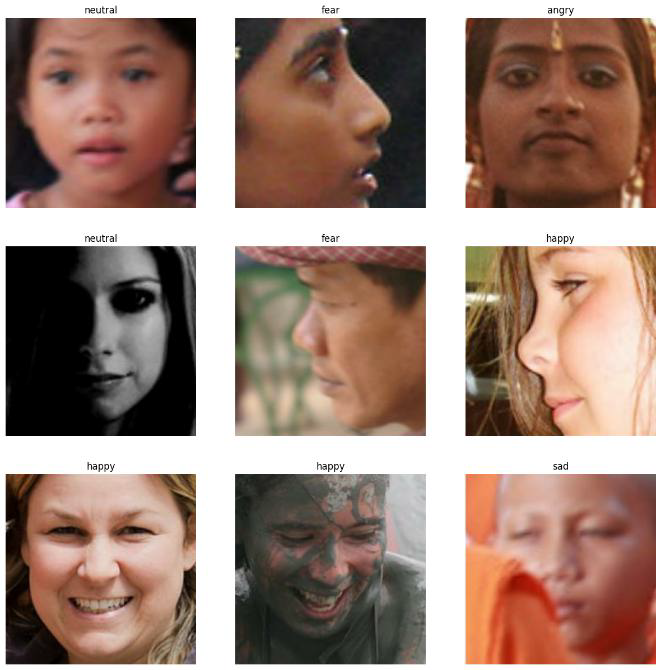}}
\caption{A Sample of the Labelled Data.}
\label{fig1}
\end{figure}

The FairFace dataset \cite{b18} was selected due to its balanced representation of race, gender, and age attributes. FairFace contains 97,698 images, each labeled with demographic attributes such as race, gender, and age. Its equitable representation of diverse demographic groups makes it an ideal choice for fairness studies in machine learning. The DeepFace pre-trained model \cite{b19} was used to generate emotion labels for the images. The emotion classes are: angry, happy, sad, fear, neutral, surprise, and disgust. This automated labelling process was applied to the entire dataset, ensuring consistent emotion annotations across training, validation, and test sets. Fig. ~\ref{fig1} presents some sample images from the labelled dataset.

\subsection{The CNN Classification Model}\label{sec3_2}
The FairFace dataset supplemented with emotion labels from DeepFace was split into training, validation, and test sets, as shown in Table \ref{tab1}. The splitting was performed in a stratified manner to maintain the proportion of gender and emotion classes in each subset. The overall gender distribution in the combined dataset is 53\% male and 47\% female.

A custom CNN model was developed for emotion classification. The architecture consists of four convolutional layers, progressively increasing filters (32, 64, 128, 256) for hierarchical feature extraction, each followed by batch normalization and max-pooling for stabilization and dimensionality reduction. The extracted 3D feature maps are flattened into a 1D vector, passed through two dense layers (256 and 128 units) with dropout for regularization, and finally outputted through a dense layer with 7 units for classification. It has 2.78M trainable parameters, efficiently balancing feature learning and prediction. Table \ref{tab_app} provides the details of the CNN architecture.

\begin{table}[hbtp]
\caption{Data Splitting}
\begin{center}
\begin{tabular}{|c|c|c|} 
 \hline
 {Data Subsets} & {Size (rows)} & {Description} \\
 \hline
 Training & 86,744 & Used to train the emotion classifier \\
 \hline
 Validation & 5,477 & Used for validation \\
 \hline
 Test & 5,477 & Used to evaluate emotion classification \\ 
 \hline
 Total & 97,698 & \multicolumn{1}{c}{}  \\ 
 \cline{1-2}
\end{tabular}
\label{tab1}
\end{center}
\end{table}

\begin{table}[hbtp]
\caption{Detailed CNN Architecture}
\begin{center}
\begin{tabular}{|p{35mm}|c|c|}
\hline
Layer (type) & Output Shape & Param \#\\ \hline
Conv2d\_9 (Conv2D) & (None, 126, 126, 32) & 896\\ \hline
batch\_normalization\_6 (BatchNormalization) & (None, 126, 126, 32) & 128\\ \hline
conv2d\_10 (Conv2D) & (None, 61, 61, 64) & 18,496\\ \hline
batch\_normalization\_7 (BatchNormalization) & (None, 61, 61, 64) & 256\\ \hline
max\_pooling2d\_6 (MaxPooling2D) & (None, 30, 30, 64) & 0\\ \hline
conv2d\_11 (Conv2D) & (None, 28, 28, 128) & 73,856\\ \hline
batch\_normalization\_8 (BatchNormalization) & (None, 28, 28, 128) & 512\\ \hline
max\_pooling2d\_7 (MaxPooling2D) & (None, 14, 14, 128) & 0\\ \hline
conv2d\_12 (Conv2D) & (None, 12, 12, 256) & 295,168\\ \hline
batch\_normalization\_9 (BatchNormalization) & (None, 12, 12, 256) & 1,024\\ \hline
max\_pooling2d\_8 (MaxPooling2D) & (None, 6, 6, 256) & 0\\ \hline
flatten\_2 (Flatten) & (None, 9216) & 0\\ \hline
dense\_6 (Dense) & (None, 256) & 2,359,552\\ \hline
batch\_normalization\_10 (BatchNormalization) & (None, 256) & 1,024\\ \hline
dropout\_4 (Dropout) & (None, 256) & 0\\ \hline
dense\_7 (Dense) & (None, 128) & 32,896\\ \hline
batch\_normalization\_11 (BatchNormalization) & (None, 128) & 512\\ \hline
dropout\_5 (Dropout) & (None, 128) & 0\\ \hline
dense\_8 (Dense) & (None, 7) & 903\\ \hline
\multicolumn{3}{|c|}{Total params: 2,785,223 (10.62 MB)}\\
\multicolumn{3}{|c|}{Trainable params: 2,783,495 (10.62 MB)}\\
\multicolumn{3}{|c|}{Non-trainable params: 1,728 (6.75 KB)}\\ \hline
\end{tabular}
\label{tab_app}
\end{center}
\end{table}

The model was trained using the Adam optimizer with a learning rate of 0.0001 and a categorical cross-entropy loss function. An early stopping mechanism was implemented, monitoring the validation loss with a patience of five epochs to prevent overfitting. The model was trained for a maximum of 10 epochs, balancing computational efficiency and performance. After training, the model's performance was evaluated on the test set. For each test sample, the model produced probabilities for the seven emotion classes, and the emotion with the highest probability was selected as the predicted label. Table \ref{tab2} shows the cross tabulation for the true emotion and the predicted emotion. The correct classifications are in boldface. The overall accuracy of the test set is 58.3\%. The test set, together with its predicted labels, will be used in the following subsection to evaluate the fairness of the CNN model.

\begin{table*}[hbtp]
\caption{Cross Tabulating True Emotion and Predicted Emotion}
\begin{center}
\begin{tabular}{|c|c|c|c|c|c|c|c|c|c|}
\hline
\multicolumn{2}{|c|}{} 
&\multicolumn{7}{c|}{Predicted Emotion from CNN Model} & \\
\cline{3-9} 
\multicolumn{2}{|c|}{}
& happy & neutral & sad & fear & angry & surprise & disgust & Total \\ 
\hline
\parbox[t]{2mm}{\multirow{7}{*}{\rotatebox[origin=c]{90}{True Emotion}}}
& happy & \textbf{1,365} & 187 & 81 & 24 & 9 & 0 & 0 & 1,666\\ \cline{2-10}
& neutral & 199 & \textbf{1,163} & 197 & 41 & 17 & 0 & 0 & 1,617\\ \cline{2-10}
& sad & 134 & 331 & \textbf{507} & 59 & 24 & 0 & 0 & 1,055\\ \cline{2-10}
& fear & 85 & 170 & 192 & \textbf{103} & 11 & 1 & 0 & 562\\ \cline{2-10}
& angry & 75 & 176 & 139 & 41 & \textbf{51} & 0 & 0 & 482\\ \cline{2-10}
& surprise & 23 & 24 & 13 & 18 & 1 & \textbf{3} & 0 & 82\\ \cline{2-10}
& disgust & 8 & 2 & 2 & 0 & 1 & 0 & \textbf{0} & 13\\ \hline
\end{tabular}
\label{tab2}
\end{center}
\end{table*}

\subsection{Bias Detection and Mitigation}\label{sec3_3}
\paragraph{Data Preparation}
Since “fear”, “angry”, “surprise” and “disgust” are small classes compared to “happy”, “neutral” and “sad”, they are group together under the label “others.” This is to ensure that there is sufficient data in each class for training the causal model for bias detection and mitigation. The test set ($n$ = 5,477) is further split into a training set (80\%, $n$ = 4,381) for causal model development and a test set (20\%, $n$ = 1,096) for the evaluation of subsequent bias mitigation. To avoid confusion, these data subsets are called the CM training set and the CM test set, respectively. “CM” here stands for “causal model.” 

To improve multiclass predicted probabilities and to establish a baseline for comparison with subsequent debiased classifications, the one-vs-all (OvA) technique \cite{b20} is applied to both the CM training set and the CM test set to calibrate the probabilities generated from the CNN model. Tables \ref{tab3} and \ref{tab4}, respectively, show the cross tabulation for the true emotion and the predicted emotion per gender for the CM training and test sets. The overall accuracies for the CM training and test sets are now, respectively, 60.6\% and 60.4\%, respectively, a slight improvement compared to 58.3\% before calibration. 

\begin{table}[hbtp]
\caption{Cross Tabulating True Emotion and Predicted Emotion (CM Training)}
\begin{center}
\begin{tabular}{|c|c|c|c|c|c|c|}
\hline
\multicolumn{2}{|c|}{} 
&\multicolumn{5}{c|}{Predicted Emotion from CNN Model} \\ 
\multicolumn{2}{|c|}{} 
&\multicolumn{5}{c|}{(Female, CM Training)} \\
\cline{3-7}
\multicolumn{2}{|c|}{}
& happy & neutral & sad & others & Total \\ \hline
\parbox[t]{2mm}{\multirow{4}{*}{\rotatebox[origin=c]{90}{True}}}
\parbox[t]{2mm}{\multirow{4}{*}{\rotatebox[origin=c]{90}{Emotion}}}
& happy & \textbf{679} & 75 & 13 & 44 & 811 \\ \cline{2-7}
& neutral & 62 & \textbf{316} & 58 & 68 & 504 \\ \cline{2-7}
& sad & 50 & 91 & \textbf{160} & 88 & 389 \\ \cline{2-7}
& others & 67 & 84 & 77 & \textbf{159} & 387 \\ \hline
\multicolumn{2}{|c|}{} 
&\multicolumn{5}{c|}{Predicted Emotion from CNN Model} \\ 
\multicolumn{2}{|c|}{} 
&\multicolumn{5}{c|}{(Male, CM Training)} \\
\cline{3-7}
\multicolumn{2}{|c|}{}
& happy & neutral & sad & others & Total \\ \hline
\parbox[t]{2mm}{\multirow{4}{*}{\rotatebox[origin=c]{90}{True}}}
\parbox[t]{2mm}{\multirow{4}{*}{\rotatebox[origin=c]{90}{Emotion}}}
& happy & \textbf{380} & 72 & 30 & 48 & 530 \\ \cline{2-7}
& neutral & 80 & \textbf{553} & 71 & 88 & 792 \\ \cline{2-7}
& sad & 42 & 126 & \textbf{180} & 114 & 462 \\ \cline{2-7}
& others & 52 & 140 & 88 & \textbf{226} & 506 \\ \hline
\end{tabular}
\label{tab3}
\end{center}
\end{table}

\begin{table}[hbtp]
\caption{Cross Tabulating True Emotion and Predicted Emotion (CM Test)}
\begin{center}
\begin{tabular}{|c|c|c|c|c|c|c|}
\hline
\multicolumn{2}{|c|}{} 
&\multicolumn{5}{c|}{Predicted Emotion from CNN Model} \\ 
\multicolumn{2}{|c|}{} 
&\multicolumn{5}{c|}{(Female, CM Test)} \\
\cline{3-7}
\multicolumn{2}{|c|}{}
& happy & neutral & sad & others & Total \\ \hline
\parbox[t]{2mm}{\multirow{4}{*}{\rotatebox[origin=c]{90}{True}}}
\parbox[t]{2mm}{\multirow{4}{*}{\rotatebox[origin=c]{90}{Emotion}}}
& happy & \textbf{168} & 10 & 5 & 10 & 193\\ \cline{2-7}
& neutral & 25 & \textbf{71} & 14 & 21 & 131\\ \cline{2-7}
& sad & 6 & 22 & \textbf{42} & 16 & 86\\ \cline{2-7}
& others & 20 & 27 & 22 & \textbf{39} & 108\\ \hline
\multicolumn{2}{|c|}{} 
&\multicolumn{5}{c|}{Predicted Emotion from CNN Model} \\ 
\multicolumn{2}{|c|}{} 
&\multicolumn{5}{c|}{(Male, CM Test)} \\
\cline{3-7}
\cline{3-7} 
\multicolumn{2}{|c|}{}
& happy & neutral & sad & others & Total \\ \hline
\parbox[t]{2mm}{\multirow{4}{*}{\rotatebox[origin=c]{90}{True}}}
\parbox[t]{2mm}{\multirow{4}{*}{\rotatebox[origin=c]{90}{Emotion}}}
& happy & \textbf{98} & 17 & 9 & 8 & 132\\ \cline{2-7}
& neutral & 9 & \textbf{136} & 22 & 23 & 190\\ \cline{2-7}
& sad & 16 & 40 & \textbf{42} & 20 & 118\\ \cline{2-7}
& others & 12 & 31 & 29 & \textbf{66} & 138\\ \hline
\end{tabular}
\label{tab4}
\end{center}
\end{table}

\begin{figure}[hbtp]
\centerline{\includegraphics[width=\linewidth]{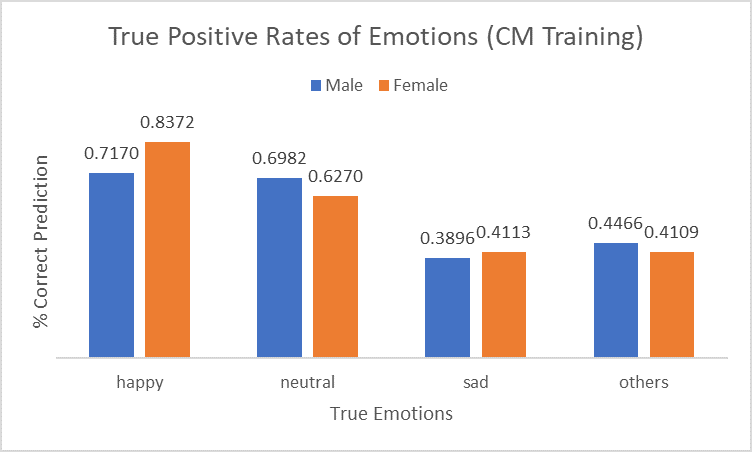}}
\caption{Gender Fairness of the CNN Model (CM Training).}
\label{fig2}
\end{figure}

\begin{figure}[hbtp]
\centerline{\includegraphics[width=\linewidth]{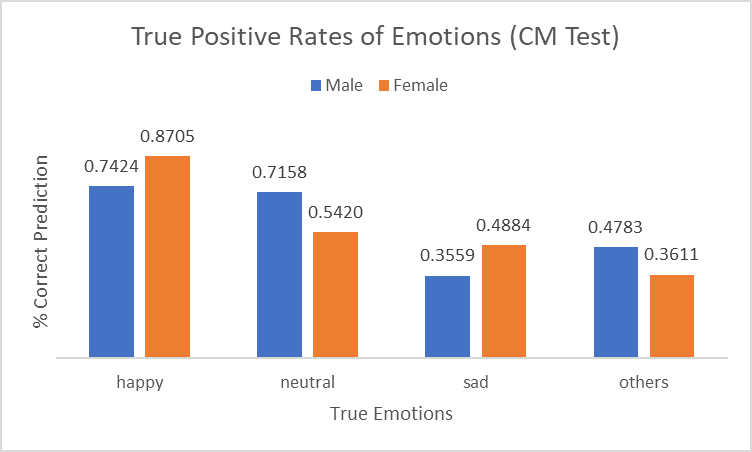}}
\caption{Gender Fairness of the CNN Model (CM Test).}
\label{fig3}
\end{figure}

\paragraph{Evaluation of Fairness of the CNN Model}
For female images, the accuracies are 62.8\% and 61.8\% for the CM training and test sets, respectively. For male images, the accuracies are 58.5\% and 59.2\% for the CM training and test sets, respectively. Graphically, Figs. ~\ref{fig2} and ~\ref{fig3} provide a more detailed comparison of the CNN model's algorithmic fairness. Both the CM training and test sets reveal the same pattern: females are more likely to be correctly classified as “happy” or “sad,” while males are more likely to be correctly classified as “neutral.” This suggests the model may find females as more emotionally expressive.

\begin{figure}[hbtp]
\centerline{\includegraphics[width=\linewidth]{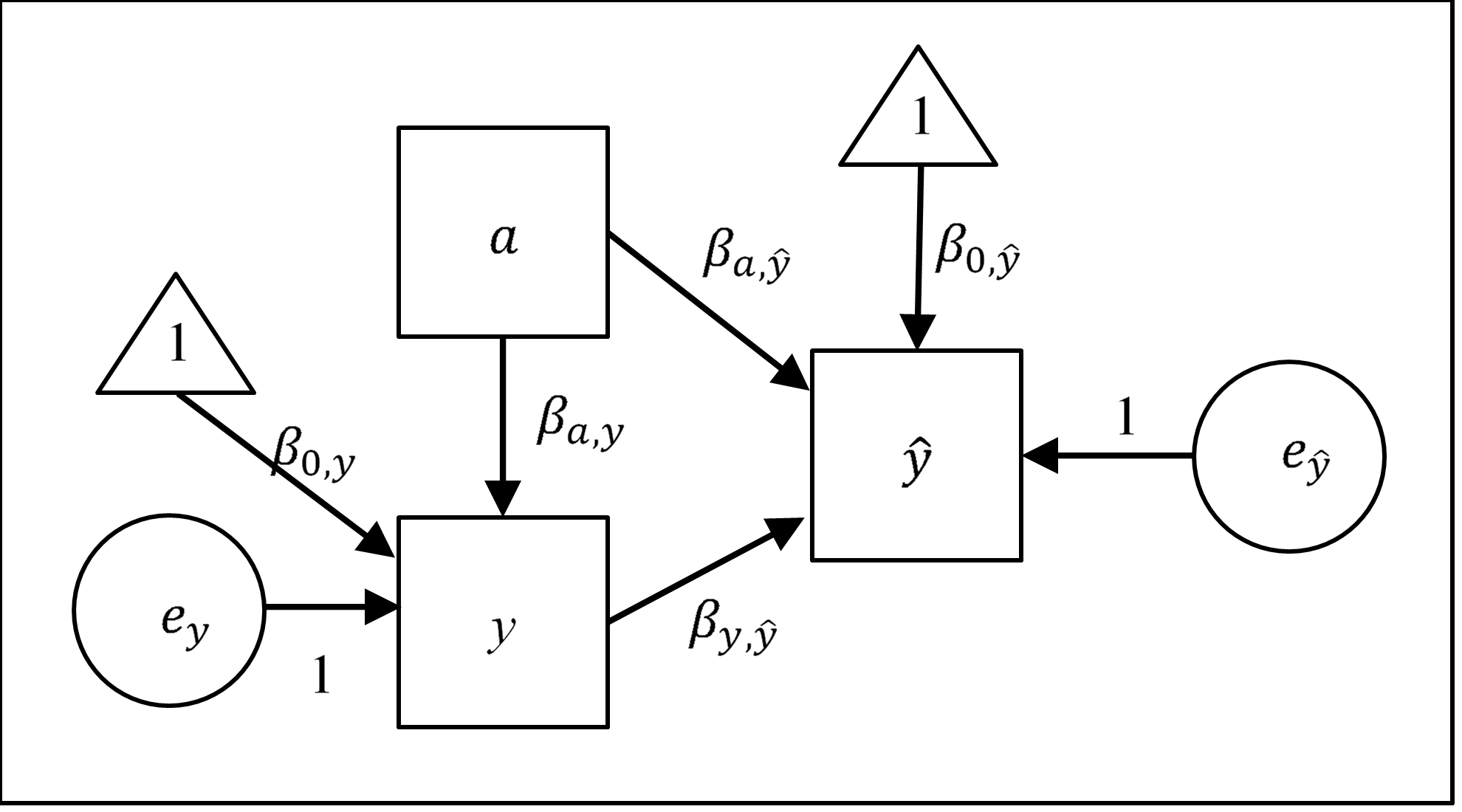}}
\caption{Causal Model for Binary Classification. \cite{b5}}
\label{fig4}
\end{figure}

\paragraph{Causal Model for Bias Detection}
Suppose we aim to reduce the accuracy difference between genders. We can apply causal modeling as proposed in \cite{b5}. Following the drawing convention in \cite{b21}, Fig. \ref{fig4} presents the causal model for binary classification. Here, \(\hat{y}\) represents the predicted probability of the binary class of interest, \(y\) is the dummy variable for that class, and \(a\) is the protected attribute (i.e., gender in this case). $e_{\hat{y}}$ and $e_y$ are simply the error terms. The $\beta$ terms are known as path coefficients and they indicate how much the dependent variables are affected by the associated predictor variables.

The causal model can be translated to the following set of equations,
\begin{eqnarray}
\label{yhat}
\hat{y} &=& \beta_{0,\hat{y}} + \beta_{a,\hat{y}} a + \beta_{y,\hat{y}} y + e_{\hat{y}} \\
\label{y}
y &=& \beta_{0,{y}} + \beta_{a,{y}} a  + e_{{y}}
\end{eqnarray}

Since we are interested only in developing a mitigation model for the predicted probabilities, we will focus only on \eqref{yhat}. To extend the method to multiclass problems, we apply the OvA technique. We build one causal model for each class, i.e.,
\[\hat{y}_{happy} = \beta_{0,\hat{y}_{happy}} + \beta_{a,\hat{y}_{happy}} a + \beta_{y,\hat{y}_{happy}} y + e_{\hat{y}_{happy}}\]
\[\hat{y}_{neutral} = \beta_{0,\hat{y}_{neutral}} + \beta_{a,\hat{y}_{neutral}} a + \beta_{y,\hat{y}_{neutral}} y + e_{\hat{y}_{neutral}}\]
\[\hat{y}_{sad} = \beta_{0,\hat{y}_{sad}} + \beta_{a,\hat{y}_{sad}} a + \beta_{y,\hat{y}_{sad}} y + e_{\hat{y}_{sad}}\]
\[\hat{y}_{others} = \beta_{0,\hat{y}_{others}} + \beta_{a,\hat{y}_{others}} a + \beta_{y,\hat{y}_{others}} y + e_{\hat{y}_{others}}\]

Since we only need to deal with one equation per emotion, we can simply use linear regression to come up with the coefficient estimates. The results are shown in Tables \ref{tab5} to \ref{tab8}. All \(p\)-values associated with “Male” are small, suggesting statistically significant gender bias. 

\begin{table}[hbtp]
\caption{Regression Results for “happy”}
\begin{center}
\begin{tabular}{|c|c|c|c|c|}
\hline
\multicolumn{2}{|c|}{Coefficient} 
& Est. & \(t\) & \(p\) \\ 
\hline
Intercept & \(\beta_{0, \hat{y}_{happy}}\) & 0.183 & 30.39 & \(<\)0.001\\ \hline
Male (\(a\)) & \(\beta_{a, \hat{y}_{happy}}\) & -0.058 & -8.013 & \(<\)0.001\\ \hline
happy (\(y_{happy}\)) & \(\beta_{y_{happy}, \hat{y}_{happy}}\) & 0.522 & 66.29 & \(<\)0.001\\ \hline
\end{tabular}
\label{tab5}
\end{center}
\end{table}

\begin{table}[hbtp]
\caption{Regression Results for “neutral”}
\begin{center}
\begin{tabular}{|c|c|c|c|c|}
\hline
\multicolumn{2}{|c|}{Coefficient} 
& Est. & \(t\) & \(p\) \\ 
\hline
Intercept & \(\beta_{0, \hat{y}_{neutral}}\) & 0.195 & 41.64 & \(<\)0.001\\ \hline
Male (\(a\)) & \(\beta_{a, \hat{y}_{neutral}}\) & 0.052 & 8.54 & \(<\)0.001\\ \hline
happy (\(y_{neutral}\)) & \(\beta_{y_{neutral}, \hat{y}_{neutral}}\) & 0.279 & 41.7 & \(<\)0.001\\ \hline
\end{tabular}
\label{tab6}
\end{center}
\end{table}

\begin{table}[hbtp]
\caption{Regression Results for “sad”}
\begin{center}
\begin{tabular}{|c|c|c|c|c|}
\hline
\multicolumn{2}{|c|}{Coefficient} 
& Est. & \(t\) & \(p\) \\ 
\hline
Intercept & \(\beta_{0, \hat{y}_{sad}}\) & 0.142 & 41.08 & \(<\)0.001\\ \hline
Male (\(a\)) & \(\beta_{a, \hat{y}_{sad}}\) & 0.025 & 5.405 & \(<\)0.001\\ \hline
happy (\(y_{sad}\)) & \(\beta_{y_{sad}, \hat{y}_{sad}}\) & 0.180 & 31.314 & \(<\)0.001\\ \hline
\end{tabular}
\label{tab7}
\end{center}
\end{table}

\begin{table}[hbtp]
\caption{Regression Results for “others”}
\begin{center}
\begin{tabular}{|c|c|c|c|c|}
\hline
\multicolumn{2}{|c|}{Coefficient} 
& Est. & \(t\) & \(p\) \\ 
\hline
Intercept & \(\beta_{0, \hat{y}_{others}}\) & 0.148 & 45.995 & \(<\)0.001\\ \hline
Male (\(a\)) & \(\beta_{a, \hat{y}_{others}}\) & 0.026 & 5.999 & \(<\)0.001\\ \hline
happy (\(y_{others}\)) & \(\beta_{y_{others}, \hat{y}_{others}}\) & 0.152 & 28.886 & \(<\)0.001\\ \hline
\end{tabular}
\label{tab8}
\end{center}
\end{table}

\paragraph{Bias Mitigation}
Following \cite{b5}, we compute the debiased probability estimates for each emotion class \(\tilde{y}\) by removing the effects of gender, i.e.,
\begin{equation}\tilde{y}_{happy} = y_{happy} + 0.058 a \label{eq7}\end{equation}
\begin{equation}\tilde{y}_{neutral} = y_{neutral} - 0.052 a \label{eq8}\end{equation}
\begin{equation}\tilde{y}_{sad} = y_{sad} - 0.025 a \label{eq9}\end{equation}
\begin{equation}\tilde{y}_{others} = y_{others} - 0.026 a \label{eq10}\end{equation}

\begin{figure}[hbtp]
\centerline{\includegraphics[width=\linewidth]{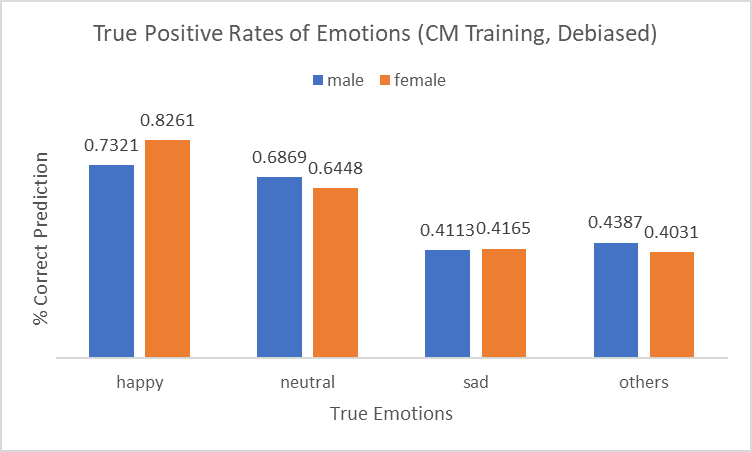}}
\caption{Gender Fairness of the CNN Model (CM Training, Debiased).}
\label{fig5}
\end{figure}

The emotion class associated with the highest debiased probability is then chosen as the predicted class. These mitigation efforts result in reduced gender gap as shown in Fig. \ref{fig5}. Compared to Fig. \ref{fig2}, we can see that the gender gaps are narrowed for all emotions.

\begin{figure}[hbtp]
\centerline{\includegraphics[width=\linewidth]{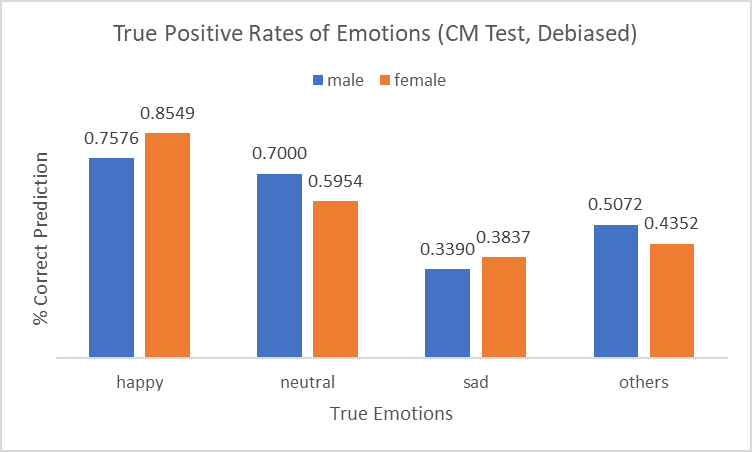}}
\caption{Gender Fairness of the CNN Model (CM Test, Debiased).}
\label{fig6}
\end{figure}

As noted previously, the beta coefficients for "Male" are statistically significant across all emotions. This is likely due to the large sample size in the training set ($n$ = 4,381). To provide a more meaningful assessment of the effectiveness of bias detection and mitigation using causal modeling, we turn to cross-validation. We again apply the mitigation models in \eqref{eq7} to \eqref{eq10} to compute the debiased probabilities and determine the predicted class using maximum probability. Fig. \ref{fig6} compares the true positive rates based on the debiased classifications for males vs females. Compared to Fig. \ref{fig3}, we again observe that the gender gaps are reduced across all emotions.

The true positive rates for some gender and emotion combinations increased, while others decreased. Overall, accuracy for the CM training set remained roughly the same, while accuracy for the CM test set improved slightly from 60.4\% to 60.8\%. For female images, the accuracies are 62.8\% and 62.4\% for the CM training set and CM test set, respectively. For male images, the accuracies are 58.7\% and 59.3\% for the CM training set and CM test set, respectively. These changes are negligible, as summarized in Table \ref{tab9}. Table \ref{tab10} compares the true positive rates for males vs. females before and after bias mitigation.

\begin{table*}[hbtp]
\caption{Accuracy Before and After Bias Mitigation}
\begin{center}
\begin{tabular}{|c|c|c|c|c|c|c|}
\hline
  & \multicolumn{3}{c|}{CM Training} &  \multicolumn{3}{c|}{CM Test}\\ \cline{2-7}
  & Before &  After & Gender Gap & Before & After & Gender Gap \\ 
  & Mitigation &  Mitigation & Reduced? & Mitigation & Mitigation & Reduced?\\ \hline
  Female & 62.8\% & 62.8\% & No & 61.8\% & 62.4\% & Yes\\ \hline
Male & 58.5\% & 58.7\% & Yes & 59.1\% & 59.3\% & Yes\\ \hline
Overall & 60.6\% & 60.6\% & No & 60.4\% & 60.8\% & Yes\\ \hline
\end{tabular}
\label{tab9}
\end{center}
\end{table*}

\begin{table*}[hbtp]
\caption{Comparing Gender Bias Before and After Bias Mitigation}
\begin{center}
\begin{tabular}{|c|c|c|c|c|c|c|c|c|}
\hline
  \multicolumn{2}{|c|}{} & \multicolumn{3}{c|}{Before Mitigation} & \multicolumn{3}{c|}{After Mitigation} & Gender Gap\\ \cline{3-8}
  \multicolumn{2}{|c|}{} & Female & Male & Gender Gap & Female & Male & Gender Gap & Reduced?\\ \hline
  CM Training & happy & 0.8372 & 0.7170 & 0.1203 & 0.8261 & 0.7321 & 0.0941 & Yes\\ \cline{2-9}
 & neutral & 0.6270 & 0.6982 & 0.0712 & 0.6448 & 0.6869 & 0.0420 & Yes\\ \cline{2-9}
 & sad & 0.4113 & 0.3896 & 0.0217 & 0.4165 & 0.4113 & 0.0052 & Yes\\ \cline{2-9}
 & others & 0.4109 & 0.4466 & 0.0358 & 0.4031 & 0.4387 & 0.0356 & Yes\\ \hline
CM Test & happy & 0.8705 & 0.7424 & 0.1280 & 0.8549 & 0.7576 & 0.0973 & Yes\\ \cline{2-9}
 & neutral & 0.5420 & 0.7158 & 0.1738 & 0.5954 & 0.7000 & 0.1046 & Yes\\ \cline{2-9}
 & sad & 0.4884 & 0.3559 & 0.1324 & 0.3837 & 0.3390 & 0.0447 & Yes\\ \cline{2-9}
 & others & 0.3611 & 0.4783 & 0.1171 & 0.4352 & 0.5072 & 0.0721 & Yes\\ \cline{2-9}
\hline
\end{tabular}
\label{tab10}
\end{center}
\end{table*}

\section{Discussion and Conclusion}
To summarize, this study investigated algorithmic bias in CNNs for emotion classification, focusing on gender as the protected attribute. The analysis leveraged the FairFace dataset, supplemented by emotional labels from the DeepFace pre-trained model. Our custom CNN model, consisting of four convolutional blocks followed by fully connected and dropout layers to mitigate overfitting, was trained using the Adam optimizer with a learning rate of 0.0001 and employed an early stopping mechanism based on validation loss. 

Causal modeling techniques were used to detect biases in the model's classifications. A path diagram was used to describe the pathways through which gender could influence the model's outputs. The path coefficients from the causal analysis revealed that gender bias significantly affected the predicted emotions. To mitigate this bias, we adjusted the class probabilities from the CNN model based on the fitted causal models and aggregated the adjusted probabilities using maximum probability. The debiased classifications reduced gender disparities across all emotions without sacrificing overall accuracy.

Our contribution to knowledge is two-fold. First, we extend the causal modeling techniques for bias mitigation to multiclass classification problems. Second, while the existing literature mostly applies post-processing of algorithmic bias to traditional machine learning algorithms, we demonstrate that it can also be effectively applied to more advanced algorithms, such as CNNs. In the future, we will further apply causal modeling to address algorithmic bias. Specifically, we will develop a stronger theoretical foundation for the conditions under which post-processing can enhance fairness while improving accuracy. We will also explore the use of causal modeling to address biases that originate from the training data.

To conclude, causal modeling is an effective approach for addressing algorithmic bias in post-processing. It is easy to implement using existing statistical tools. Furthermore, its statistical roots ensure ease of interpretation. When used to complement a complex AI model, a causal model can enhance explainability, enabling stakeholders to better understand the nature of the bias and how it can be mitigated.

\section*{Acknowledgment}
ChatGPT was used to review the grammar of this paper.

\end{document}